\documentclass[10pt,twocolumn,letterpaper]{article}

\usepackage{iccv}
\usepackage{times}
\usepackage{epsfig}
\usepackage{graphicx}
\usepackage{amsmath}
\usepackage{amssymb}
\usepackage{amsfonts}
\usepackage{bm}

\usepackage{booktabs}
\usepackage{multirow}

\usepackage[accsupp]{axessibility}  

\usepackage[pagebackref=true,breaklinks=true,letterpaper=true,colorlinks,bookmarks=false]{hyperref}

\usepackage[capitalize]{cleveref}
\crefname{section}{Sec.}{Secs.}
\Crefname{section}{Section}{Sections}
\Crefname{table}{Table}{Tables}
\crefname{table}{Tab.}{Tabs.}

\iccvfinalcopy 

\def\iccvPaperID{5385} 

\ificcvfinal\fi

\begin{document}

\title{Interaction-aware Joint Attention Estimation Using People Attributes}


\author{Chihiro Nakatani$^{1}$\quad Hiroaki Kawashima$^{2}$\quad Norimichi Ukita$^{1}$\\
$^{1}$ Toyota Technological Institute, Japan \quad $^{2}$ University of Hyogo, Japan}

\maketitle
\ificcvfinal\thispagestyle{empty}\fi

\begin{abstract}
This paper proposes joint attention estimation in a single image.
Different from related work in which only the gaze-related attributes of people are independently employed, (i) their locations and actions are also employed as contextual cues for weighting their attributes, and (ii) interactions among all of these attributes are explicitly modeled in our method.
%
For the interaction modeling, we propose a novel Transformer-based attention network to encode joint attention as low-dimensional features.
We introduce a specialized MLP head with positional embedding to the Transformer so that it predicts pixelwise confidence of joint attention for generating the confidence heatmap.
This pixelwise prediction improves the heatmap accuracy by avoiding the ill-posed problem in which the high-dimensional heatmap is predicted from the low-dimensional features.
The estimated joint attention is further improved by being integrated with general image-based attention estimation.
Our method outperforms SOTA methods quantitatively in comparative experiments.
Code: \url{https://github.com/chihina/PJAE}.
\end{abstract}



\section{Introduction}
\label{section:introduction}

Attention analysis 
enables various applications, such as 
customer's interest estimation~\cite{bib:e_see_track},
analyzing atypical gaze perception in autism spectrum disorder~\cite{bib:autism_iccv2017,bib:autism_eccv2019}, and
human-robot interaction~\cite{bib:human_robot_iros_2018}.
While attention is represented as a point, region, or object in the literature, we represent it as an attention point, AP, because a point can be used in any applications as an elemental representation.
The confidence distribution of APs can be expressed in a heatmap image~\cite{bib:detecting_attended,bib:muggle,bib:inferring_shared_attention,DBLP:conf/cvpr/TuMDGZS22} where each pixel value represents the confidence.

Attention estimation has two categories: single attention estimation~\cite{bib:gaze_follow,bib:gaze_follow_video,bib:connecting_gaze,bib:gaze_multiple_cues,bib:detecting_attended,bib:dual_attention,DBLP:conf/cvpr/TuMDGZS22,Hu_2023_CVPR} and joint attention estimation~\cite{bib:inferring_shared_attention,bib:muggle,bib:social_saliency_field,bib:social_saliency_prediction,Horanyi_2023_CVPR}.
While single attention estimation targets the attention of a person,
attention shared by multiple people is detected by joint attention estimation.

In single attention estimation, an AP is estimated based on the gaze direction of a target person~\cite{DBLP:conf/eccv/ParkSH18,DBLP:journals/pami/LiuYMO21,DBLP:conf/eccv/FischerCD18,DBLP:conf/cvpr/MurthyB21,DBLP:conf/iccv/LeeYO15,DBLP:conf/iccv/KuhnkeO19,DBLP:journals/pami/GengQHZ22,DBLP:conf/cvpr/KothariMIBPK21} and the saliency map of a scene image~\cite{DBLP:conf/cvpr/LeeTK16,DBLP:conf/cvpr/OhBKAFS17,DBLP:conf/iccv/KellnhoferRSM019,DBLP:conf/iccv/YuZLZ19,DBLP:conf/bmvc/ZhouG20,DBLP:conf/cvpr/KothariMIBPK21} in general.
By simply aggregating (e.g., averaging) multi-people APs that are independently estimated by single attention estimation, a joint AP can be estimated~\cite{bib:detecting_attended,DBLP:conf/cvpr/TuMDGZS22}.
However, the APs of multiple people are not independent but jointly correlated in context.

\begin{figure}[t]
\begin{center}
\includegraphics[width=\columnwidth]{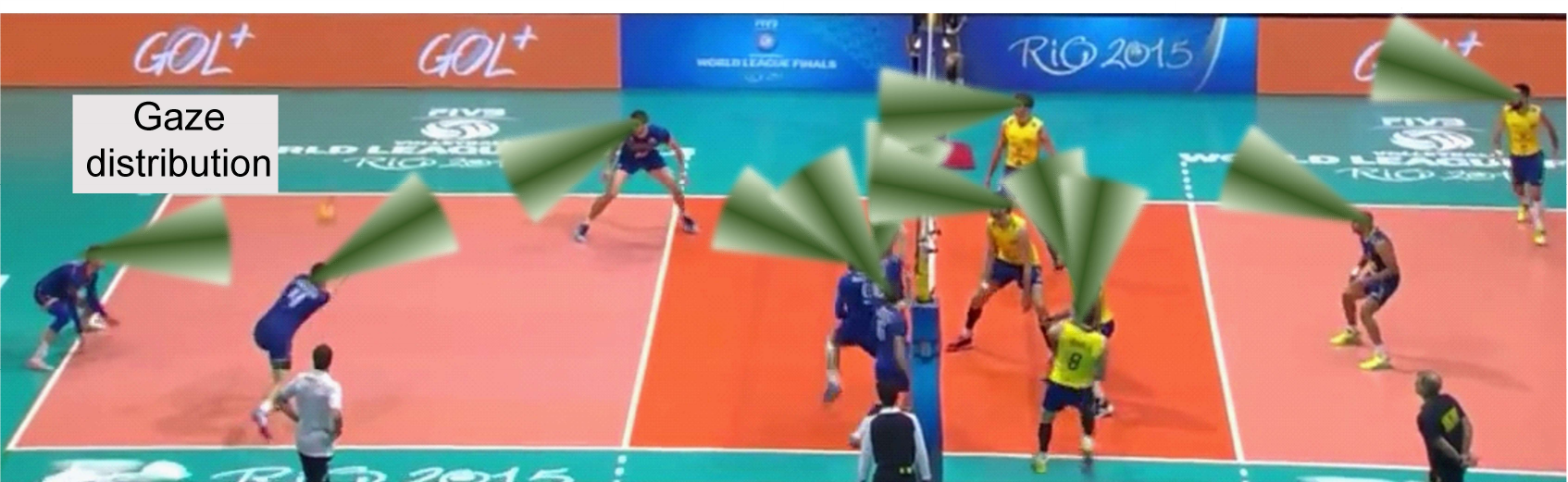}\\
{\small (a) Previous methods~\cite{bib:detecting_attended,bib:muggle,bib:inferring_shared_attention,DBLP:conf/cvpr/TuMDGZS22}}
\includegraphics[width=\columnwidth]{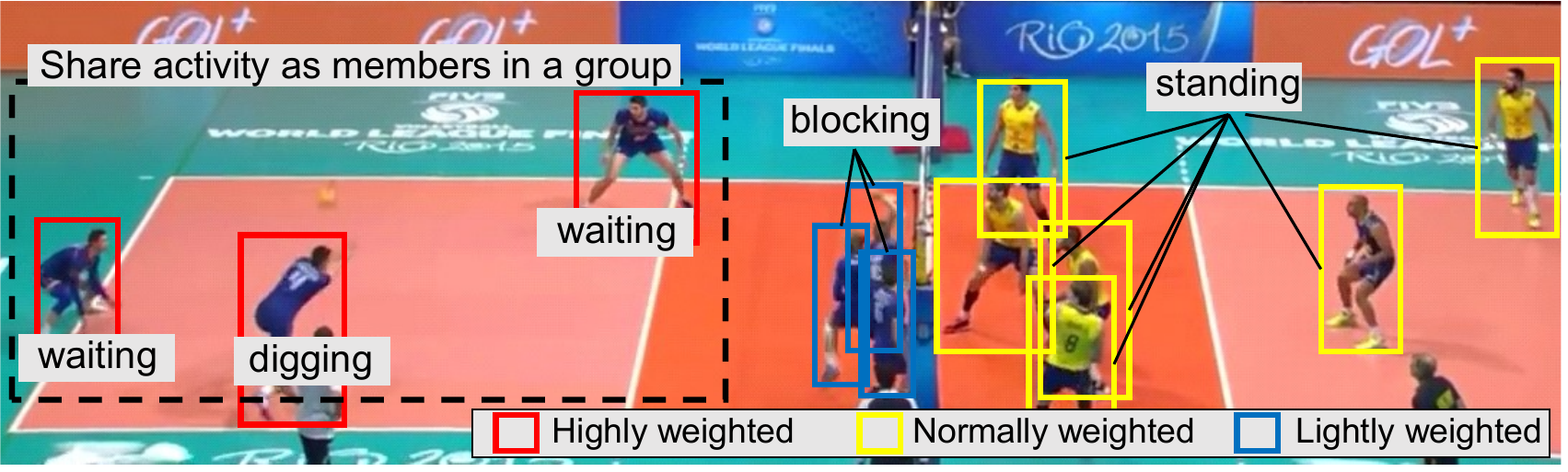}\\
{\small (b) Ours: gaze distributions are omitted for simple visualization}
\end{center}
\caption{Difference between joint attention estimation methods.
(a) Aggregating the gaze-related features of equally-weighted people without considering their interaction.
(b) Aggregating the attributes of people weighted by their contextual attributes (i.e., locations and actions) via the interaction among all the attributes.}
\label{fig:top}
\end{figure}

For joint attention estimation with such contextual correlation, in~\cite{bib:inferring_shared_attention,bib:muggle}, only the gaze-related attributes of people (e.g., ``Gaze distribution'' in Fig.~\ref{fig:top} (a)) are employed.
Such a straightforward approach has the problems below:
\begin{itemize}
\item {\bf No contribution weights of people:}
Some people share attention, but others do not.
The latter people should not affect joint attention estimation, while all people are equally weighted in~\cite{bib:inferring_shared_attention,bib:muggle}.
\item {\bf No explicit interaction among people attributes:} 
As contextual cues related to joint attention, not only gazes~\cite{bib:inferring_shared_attention,bib:muggle} but also other attributes of people, such as their locations and actions, are useful.
Their interactions are also informative.
For example, nearby people doing the same action may share the AP.
Such interactions among people attributes are neglected in~\cite{bib:inferring_shared_attention,bib:muggle}.
\end{itemize}

These problems are resolved by the following novel contributions in this paper, as illustrated in Fig.~\ref{fig:top} (b):
\begin{itemize}
\item {\bf Activity awareness:} 
Each person's activity such as the location and action can be an important clue for joint attention estimation.
For example, people sharing the AP tend to share their activities as group members.
This assumption motivates us to focus on what and where each person is doing, namely the location and action of each person, to weight the contributions of people for joint attention estimation.
\item {\bf Interaction awareness:} 
Interactions among people attributes are explicitly modeled by our Position-embedded Joint Attention Transformer (PJAT), where a self-attention mechanism extracts the features of people sharing the AP.
\item {\bf Pixelwise joint attention heatmapping:} 
While the extracted joint-attention features are efficiently but sufficiently low-dimensional, it is ill-posed to estimate a high-dimensional heatmap image representing the AP confidences from such low-dimensional features.
To avoid such an ill-posed estimation problem, we employ a network with image-coordinate embedding for estimating the AP confidence pixelwise. 
\end{itemize}


\section{Related Work}
\label{section:related}

\subsection{Single Attention Estimation}
\label{subsection:single_attention}

To understand a person's attention in a scene, appearance cues observed in the person's head, face, and eye images are important.
Scene image features are also useful to extract saliency that attracts people’s attention.
Recasens~\textit{et al}.~\cite{bib:gaze_follow,bib:gaze_follow_video} and Chong~\textit{et al}.~\cite{bib:connecting_gaze} fuse CNN features extracted from a whole image and a cropped face image.
Chong \textit{et al}.~\cite{bib:detecting_attended} employ LSTM~\cite{bib:lstm} for fusing these two kinds of features extracted in a video. Tu~\textit{et al}.~\cite{DBLP:conf/cvpr/TuMDGZS22} simultaneously estimate heads and their APs from a whole image.

Rather than the raw images of the head, face, and eyes used in the aforementioned methods, the gaze direction estimated from these images is more informative
for identifying attention.
Since the estimated gaze direction is not accurate enough, it is in general extended to a more noise-robust representation, such as a fan shape expressing the probabilistic distribution of the gaze direction~\cite{DBLP:conf/accv/LianYG18,bib:dual_attention,li2021iccv}.
As the scene features, features extracted from each object region can be more useful than those in the whole image~\cite{bib:gaze_multiple_cues}.

\subsection{Joint Attention Estimation}
\label{subsection:multi_attention}

Joint attention estimation merges the APs of multiple people.
For such 
estimation, gaze maps are superimposed to yield a social saliency field whose modes are regarded as multiple joint APs in~\cite{DBLP:conf/nips/ParkJS12}.
In~\cite{bib:social_saliency_field}, the spatial relationship between multiple gaze directions and their attention is modeled via latent social charges inspired by Coulomb's law.
As well as single attention estimation, joint attention estimation can be achieved by both raw head, face, and eye images~\cite{DBLP:conf/wacv/SumerGTK20} and gaze directions estimated in these images~\cite{bib:inferring_shared_attention,bib:muggle}.
For example, in~\cite{bib:inferring_shared_attention}, a fixed-size fan-shaped gaze map is drawn from the gaze direction of each person, and a CNN fuses the averaged gaze map of all people and the region proposal map of objects (i.e., saliency map)~\cite{bib:edge_boxes}.
In~\cite{bib:muggle}, LSTM fuses the gaze maps
observed in an image.
Such CNN and LSTM might weight the gaze maps of people for joint attention estimation.
However, the fixedly-shaped gaze maps cannot represent more flexible weights determined by interactions among people attributes (e.g., their locations, gaze directions, and actions). Such flexible weights are estimated by a self-attention mechanism in our method.

\subsection{Location- and Action-, and their Interaction-Awareness}

Our method is aware not only of the gaze directions of people but also of their other attributes such as the locations and actions, which are informative for a variety of tasks as follows.
The locations of people provide a meaningful context for individual reasoning~\cite{DBLP:conf/cvpr/GallagherC09}, person re-identification~\cite{DBLP:journals/cviu/UkitaMH16}, 
people grouping~\cite{DBLP:conf/fgr/SchmuckC21}, and
trajectory prediction~\cite{DBLP:conf/cvpr/SuHSP17}.
The action class of each person is informative for human pose estimation~\cite{DBLP:conf/eccv/GallYG10,DBLP:conf/fgr/IqbalGG17}, human motion synthesis~\cite{DBLP:journals/corr/abs-2104-05670,men2021cg}, action anticipation~\cite{DBLP:conf/iccv/AkbarianSSFPA17}, human-human interaction estimation~\cite{DBLP:conf/cvpr/ZhongQDT21}, human-object interaction estimation~\cite{DBLP:conf/ijcai/LinZX20},
and group activity recognition~\cite{bib:groupformer,DBLP:conf/cvpr/WuWWGW19,bib:crm_gar,DBLP:conf/cvpr/GavrilyukSJS20,DBLP:conf/mva/NakataniSU21}.
We can also simultaneously take into account the locations and the actions for further improving individual and group activity recognition~\cite{DBLP:conf/eccv/EhsanpourASSRR20}, while these attributes are just implicitly encoded by general image feature extraction in~\cite{DBLP:conf/eccv/EhsanpourASSRR20}.

While the effects of the locations and actions of people are validated for the above tasks, their effectiveness is
unclear for joint attention estimation.
For example, in~\cite{bib:inferring_shared_attention,bib:muggle}, the location is not directly used for attention estimation, while it is used for representing the gaze distribution, as shown in Fig.~\ref{fig:top} (a).
Our novelty in PJAT lies not in just verifying their effectiveness but in how to model interactions
among these attributes (i.e., interaction awareness).


\section{Proposed Method}

\begin{figure*}[t]
\begin{center}
\includegraphics[width=\textwidth]{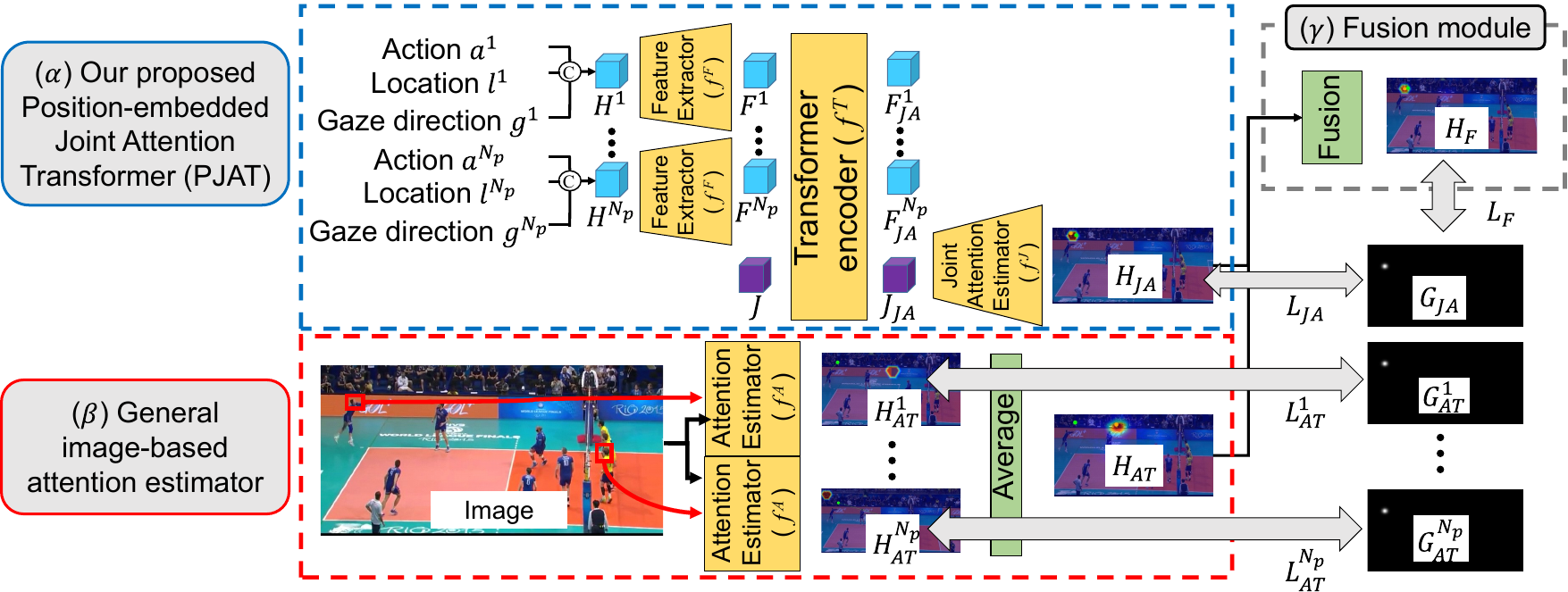}\\
\end{center}
\caption{Overview of the proposed method consisting of the three modules. ($\alpha$) Our proposed Position-embedded Joint Attention Transformer (PJAT) for joint attention estimation. In PJAT, a Transformer encoder models interactions among the attributes of people. ($\beta$) An image-based attention estimator. Any general attention estimation method can be fused with PJAT for improvement. ($\gamma$) Fusion module. The heatmaps obtained from ($\alpha$) and ($\beta$), denoted respectively by $H_{JA}$ and $H_{AT}$, are fused into the final heatmap, $H_{F}$.}
\label{fig:proposed_method}
\end{figure*}

The overview of the proposed method, consisting of three modules ($\alpha$), ($\beta$), and ($\gamma$), is illustrated in Fig.~\ref{fig:proposed_method}.
The attributes of each person are extracted from an image (Sec.~\ref{subsection:pre_process}).
The extracted people attributes are fed into the Transformer encoder in ($\alpha$) PJAT for interaction-aware joint attention estimation (Sec.~\ref{subsection:transformer} and Sec.~\ref{subsection:joint_attention}).
The estimated joint attention
is integrated with the one estimated by ($\beta$) a general image-based network for further improvement in ($\gamma$) the fusion module (Sec.~\ref{subsection:fusion}).

\subsection{Pre-processes for Person Attributes}
\label{subsection:pre_process}

The following pre-processes are used in our implementation, while these pre-processes are modularized so that they can be easily replaced with any SOTA methods.

\noindent{\bf Location detection.}
Our method employs the location of each person (denoted by $\bm{l}$) as one of the person attributes.
While any location in the body can be regarded as the person's  location, we use the head position as the person's location because the head is the edge point of a gaze line.
For this head detection, the pretrained YOLOv5~\cite{bib:yolov5} is finetuned with the head bounding boxes in each dataset.

\noindent{\bf Gaze direction estimation.}
The head bounding box of each person
is fed into the gaze direction estimator.
This estimator is a simple network consisting of VGG-16 for feature extraction followed by two fully-connected layers with output sizes of 64 and 2, in accordance with~\cite{bib:muggle,bib:inferring_shared_attention}.
The last output is a vector consisting of $x$ and $y$ directions (denoted by $\bm{g} = (g_{x}, g_{y})$) whose norm is normalized to one.

\noindent{\bf Action recognition.}
As with the head, a full-body bounding box is detected by YOLOv5~\cite{bib:yolov5}.
This bounding box is fed into an action recognition network (ARG~\cite{DBLP:conf/cvpr/WuWWGW19} in our experiments).
Given $N_{a}$ action classes, this network outputs a $N_{a}$-dimensional probability vector (denoted by $\bm{a}$) in which $j$-th component is the probability of $j$-th action class.


\subsection{Transformer Encoder for Feature Interaction}
\label{subsection:transformer}

\begin{figure}[t]
\begin{center}
\includegraphics[width=\columnwidth]{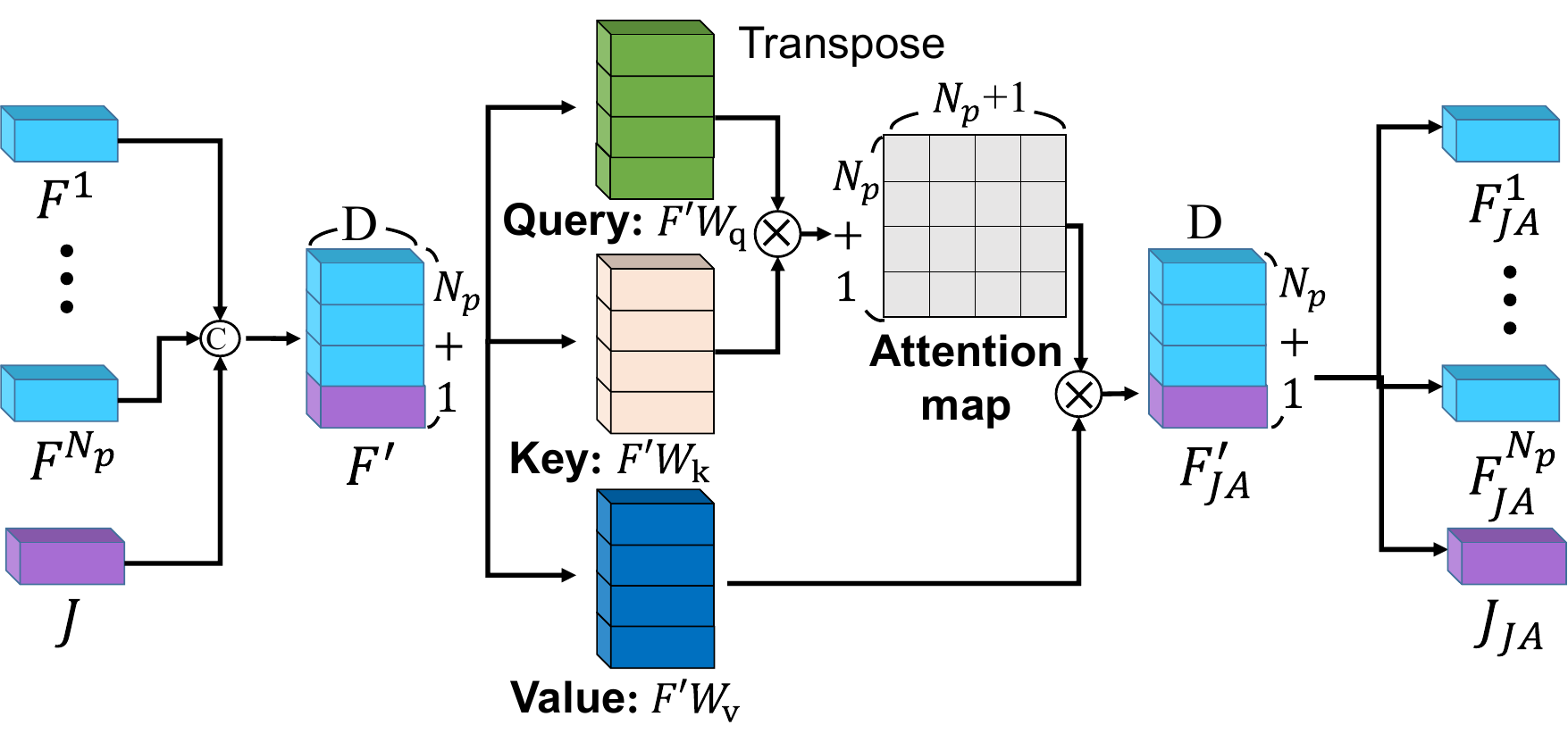}
\end{center}
\vspace*{-1mm}
\caption{Self-attention network in the Transformer encoder.
Self-attention models interactions among people attributes. The interactions are embedded into the joint attention feature $\bm{J}_{JA}$ using the features of individual people (i.e., $\bm{F}^{1}, \bm{F}^{2}, \cdots, \bm{F}^{N_{p}}$) and a learnable joint attention token (i.e., $\bm{J}$).}
\label{fig:self_attention}
\end{figure}

The interaction among the features of people (i.e., location $\bm{l}^{i}$, gaze direction $\bm{g}^{i}$, and action $\bm{a}^{i}$, where $i$ denotes the ID of each person), extracted by the pre-processes described in Sec.~\ref{subsection:pre_process}, is the main focus of this paper.
We here employ Transformer to model such interactions.
Transformer~\cite{bib:transformer} has been proven in various fields to be powerful for modeling interactions of entities 
(e.g., spatial interaction among image patches split from an image for vision tasks~\cite{DBLP:conf/iclr/DosovitskiyB0WZ21,DBLP:conf/iccv/LiuL00W0LG21} and people interaction for group activity recognition~\cite{DBLP:conf/cvpr/0002Z0Y0CQ22,DBLP:conf/cvpr/GavrilyukSJS20}).
With the self-attention mechanism, Transformer successfully handles the interaction of multiple people. This mechanism is expected to play a crucial role in joint attention estimation because it can directly reason about who shares attention by using each person's location, gaze direction, and action. 
In addition, the characteristics of (i) accepting variable-length input and (ii) its permutation-invariant property are particularly important for our joint attention estimation problem, where the number and the order of detected people may change between images due to imperfect human detection.

The attributes of $i$-th person are concatenated to be $\bm{H}^{i}=(\bm{l}^{i}, \bm{a}^{i}, \bm{g}^{i})$.
$\bm{H}^{i}$ is fed into the feature extractor ($f^{F}$ in Fig.~\ref{fig:proposed_method}) consisting of two fully-connected layers.

The features of all $N_{p}$ people, each of which is a $D$-dimensional vector (denoted by $\bm{F}^{i}$), are fed into the Transformer encoder ($f^{T}$ in Fig.~\ref{fig:proposed_method}).
$f^{T}$ consists of two transformer encoder layers with multi-head attention.
Each transformer encoder is composed of the self-attention network, feed forward layer, layer normalization, and residual path, as with~\cite{bib:transformer}.
In the self-attention network shown in Fig.~\ref{fig:self_attention}, all the features and a learnable token~\cite{DBLP:conf/iclr/DosovitskiyB0WZ21,DBLP:conf/iccv/LiZWYYXZ21} of joint attention (denoted by $\bm{J}$) are concatenated to be $F'$.
$F'$ is used to compute a query matrix $Q = F^{'} W_{q}$, a key matrix $K = F' W_{k}$, and a value matrix $V = F' W_{v}$, where $W_{q}$, $W_{k}$, and $W_{v}$ denote $D \times D$ learnable weight matrices. 
With $Q$, $K$, and $V$, $F'_{JA}$ is defined to be $softmax \left( \frac{Q K^{T}}{\sqrt{D}} \right) V$.

$F'_{JA}$ is split into the feature vectors of $N_{p}$ people (denoted by $\bm{F}^{1}_{JA}, \cdots, \bm{F}^{N_{p}}_{JA}$) and the joint attention feature $\bm{J}_{JA}$.
This self-attention mechanism allows us to optimize $\bm{J}_{JA}$ by taking into account the mutual interaction between the people attributes (i.e., $\bm{l}^{i}$, $\bm{a}^{i}$, and $\bm{g}^{i}$ of all $N_{p}$ people).


\begin{figure}[t]
\begin{center}
\includegraphics[width=\columnwidth]{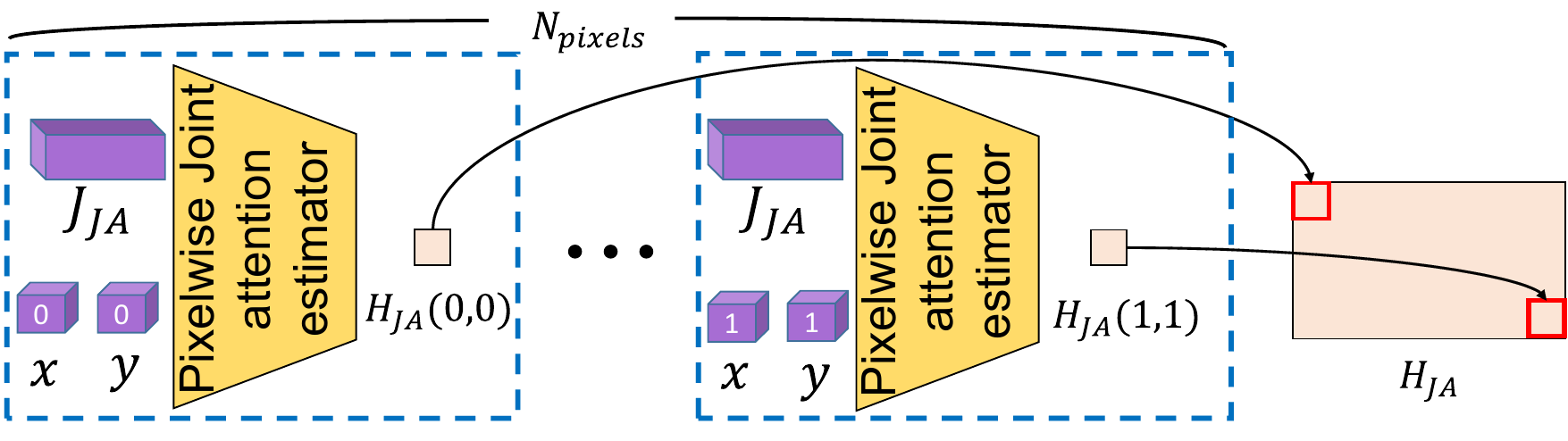}
\end{center}
\caption{Pixelwise joint attention estimation network. Coordinates of each pixel (denoted by $x$ and $y$) and extracted joint attention feature $J_{JA}$ are fed into the network in which joint attention probability (denoted by $H_{JA}(x,y)$) is estimated. 
}
\label{fig:ja_estimator}
\end{figure}

\subsection{Pixelwise Joint Attention Estimator}
\label{subsection:joint_attention}

With $\bm{J}_{JA}$, a
low-dimensional latent vector, PJAT aims to generate a high-dimensional heatmap whose a value in each pixel $(x, y)$ is the probability of joint attention in $(x, y)$.
While such a map can be directly estimated from $\bm{J}_{JA}$ using fully-connected layers~\cite{DBLP:conf/iccv/LiZWYYXZ21}, it is difficult to yield a map robustly due to its ill-posed nature.
We therefore propose pixelwise estimation of a heatmap while maintaining the spatial relationship between pixels by positional embeddings.

The proposed joint attention estimator ($f^{J}$ in Fig.~\ref{fig:proposed_method}) consists of three fully-connected layers followed by the Sigmoid activation at the last layer, as shown in Fig.~\ref{fig:ja_estimator}.
To estimate the probability of joint attention at $(x, y)$, we feed the position $(x, y)$ together with the encoded vector $\bm{J}_{JA}$ to the estimator by concatenating $(x, y)$ with $\bm{J}_{JA}$.
Its output is obtained between $0$ and $1$ via the Sigmoid activation, which is regarded as the probability value (denoted by $H_{JA} (x_{i}, y_{i})$).
We attach this specialized head to the Transformer encoder introduced in Sec.~\ref{subsection:transformer} to constitute PJAT.


\subsection{Fusion with Image-based Attention Estimation}
\label{subsection:fusion}

\noindent{\bf Scene features.}
As described in Sec.~\ref{subsection:transformer} and Sec.~\ref{subsection:joint_attention}, PJAT estimates joint attention by focusing on interactions among people attributes.
However, there are some specific situations where the people attributes alone cannot estimate joint attention well.
For example, when the number of people present in a scene is small, 
precise estimation might be difficult due to their sparse gaze distributions.
A larger number of people in a scene does not necessarily improve the estimation accuracy; for example, when the gaze directions are almost parallel, their intersection is sensitive to their noise.
Here, using scene features extracted from each image may alleviate such difficulties. For example, visual saliency is demonstrated as useful appearance information in previous attention estimation methods~\cite{bib:detecting_attended,DBLP:conf/cvpr/TuMDGZS22}. 
That is, the distribution of saliency captured by scene features helps localize the AP of each person in the gaze direction.
While the AP of each person is independently estimated in single attention estimation~\cite{bib:detecting_attended,DBLP:conf/cvpr/TuMDGZS22}, the image-based scene features also benefit joint attention estimation~\cite{bib:inferring_shared_attention} regardless of the number of people in a scene.

\noindent{\bf Fusion.}
Based on the discussion above, we design our method to fuse ``the joint attention heatmaps $H_{JA}$ estimated by PJAT'' and ``a map $H_{AT}$ estimated by an image-based attention estimator ($f^{A}$ in Fig.~\ref{fig:proposed_method})'' into the final joint attention heatmap $H_{F}$.
We here employ DAVT~\cite{bib:detecting_attended}, a SOTA attention estimation method using image-based scene features, as $f^{A}$.
In DAVT, the average of individual attention heatmaps is regarded as $H_{AT}$. $H_{F}$ is computed by the weighted fusion as follows: $H_{F}=W_{JA}H_{JA}+W_{AT}H_{AT}$, where $W_{JA}$ and $W_{AT}$ are weight coefficients. By training these coefficients for each dataset, our network fuses two heatmaps based on the scene properties.

\noindent{\bf Training objective.}
The overall network consisting of PJAT, DAVT, and the fusion module is trained with the heatmap estimation loss $L_{ALL}=L_{JA}+L_{AT}+L_{F}$, where $L_{JA}$, $L_{AT}$, and $L_{F}$ denote the loss functions used for training PJAT, DAVT, and the fusion module, respectively.
All the loss functions are based on the mean squared errors:
$L_{JA} = \sum_{n}(H_{JA}^{n}-G_{JA}^{n})^2$, $L_{AT} = \frac{1}{N_p} \sum_{i}\sum_{n}(H_{AT}^{i,n}-G_{AT}^{i,n})^2$, and $L_{F} = \sum_{n}(H_{F}^{n}-G_{JA}^{n})^2$, where $H_{JA}^{n}$, $H_{F}^{n}$, and $G_{JA}^{n}$ denote the $n$-th pixel value of the heatmaps estimated by PJAT, estimated by the fusion module, and given by the ground-truth, respectively.
$H_{AT}^{i,n}$ and $G_{AT}^{i,n}$ denote the $n$-th pixel value of the $i$-th person's heatmaps estimated by the image-based attention estimator and given by the ground-truth, respectively.
$G_{JA}$ and $G_{AT}^{i}$ are generated by drawing the 2D Gaussian distribution so that its center is located at the given ground-truth position of joint attention and $i$-th person's attention, respectively.


\section{Experiments}
\label{section:experiments}

\begin{table}[t]
    \centering
    \caption{
    Experimental conditions on the Volleyball (Vol) and VideoCoAtt (Vid) datasets. People attributes given by prediction (Pr) and ground-truth (GT) are used in Ex.1 and Ex.2, respectively.
    The condition details are described in the supplementary material.}
    \label{table:experiments}
    \vspace*{2mm}
    \begin{tabular}{l|l||c||l|l|l} \hline
    \multicolumn{2}{c|}{} & Att & Body & Head & Action \\ \hline \hline
    \multirow{2}{*}{Vol} & Ex.1 & $\bm{l}$, $\bm{g}$, $\bm{a}$ & Pr & Pr in image & Pr\\ \cline{2-6}
    & Ex.2 & $\bm{l}$, $\bm{g}$, $\bm{a}$ & GT & Pr in GT body & GT \\ \hline
    \multirow{2}{*}{Vid} & Ex.1 & $\bm{l}$, $\bm{g}$ & -- & Pr in image & --\\ \cline{2-6}
    & Ex.2 & $\bm{l}$, $\bm{g}$ & -- & GT & -- \\ \hline
    \end{tabular}
\end{table}

\begin{table*}[t]
    \centering
    \caption{Quantitative comparison on the Volleyball dataset evaluated in the two experimental conditions mentioned in Sec.~\ref{subsection:exp_datasets}. Results obtained in ball detection, Ex.1, and Ex.2 are separated by double lines. Dist: the mean distance between the ground-truth and estimated joint APs. Thr: the threshold for the joint AP detection rate. The best result in each column is colored in \textcolor{red}{red}.}
    \begin{tabular}{l|r|r|r|r|r|r} \hline
    Method & Dist ($x$) $\downarrow$ & Dist ($y$) $\downarrow$ & Dist $\downarrow$ & Thr=30 $\uparrow$ & Thr=60 $\uparrow$ & Thr=90 $\uparrow$ \\ \hline \hline
    Ball detection~\cite{bib:centernet} & 147.6 & 66.5 & 174.8 & 54.3 & 56.0 & 58.5 \\ \hline \hline
    ISA~\cite{bib:inferring_shared_attention} (Ex.1) & 53.1 & 35.5 & 70.1 & 60.7 & 69.7 & 75.9 \\ \hline
    DAVT~\cite{bib:detecting_attended} (Ex.1) & 60.2 & 28.1 & 72.0 & 62.0 & 72.8 & 78.6 \\ \hline
    Ours (Ex.1) & \textcolor{red}{44.1} & \textcolor{red}{25.2} & \textcolor{red}{56.0} & \textcolor{red}{64.5} & \textcolor{red}{76.8} & \textcolor{red}{83.0} \\ \hline \hline
    ISA~\cite{bib:inferring_shared_attention} (Ex.2) & 36.7 & 24.7 & 48.7 & 46.0 & 79.1 & 92.8 \\ \hline
    DAVT~\cite{bib:detecting_attended} (Ex.2) & 65.2 & 29.7 & 77.4 & 59.7 & 69.7 & 76.6 \\ \hline
    Ours (Ex.2) & \textcolor{red}{9.3} & \textcolor{red}{4.7} & \textcolor{red}{11.4} & \textcolor{red}{96.3} & \textcolor{red}{98.9} & \textcolor{red}{99.6} \\ \hline
    \end{tabular}
    \label{table:comparison_volleyball}
\end{table*}

\begin{figure*}[t]
  \begin{center}
  \includegraphics[width=\textwidth]{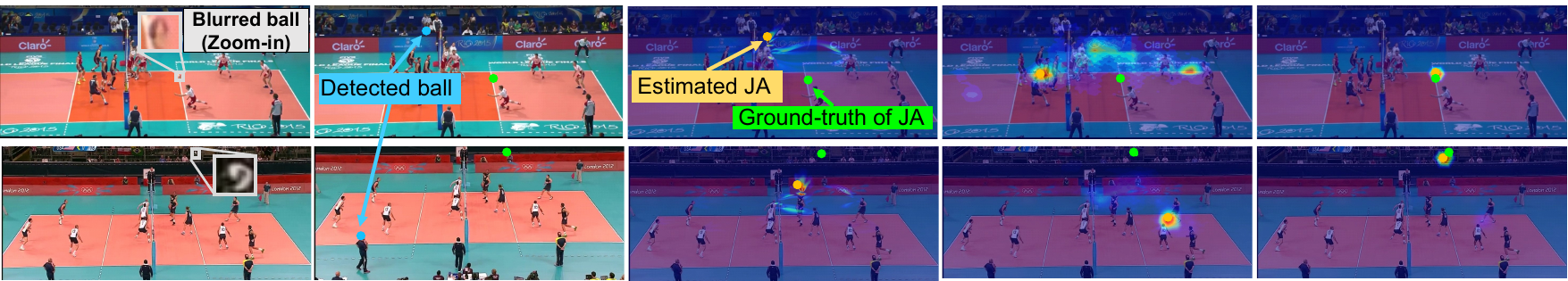}
  Image\hspace*{18mm}
  Ball detection~\cite{bib:centernet}\hspace*{20mm}
  ISA~\cite{bib:inferring_shared_attention}\hspace{22mm}
  DAVT~\cite{bib:detecting_attended}\hspace{20mm}
  Ours~\hspace*{3mm}
  \end{center}
  \caption{Visual comparison on the Volleyball dataset. Estimated joint attention heatmap is overlaid on the image. Green and yellow circles indicate the ground-truth and estimated joint APs, respectively.}
  \label{fig:comparison_volley_previous}
\end{figure*}

\begin{table*}[t]
    \centering
    \caption{Quantitative comparison on the VideoCoAtt dataset. Accuracy and F-score: metrics for the joint AP prediction on the threshold given by validation data. AUC: area under the ROC curve for the joint AP prediction.}
    \begin{tabular}{l|r|r|r|r|r|r|r|r|r} \hline
    Method & Dist ($x$) & Dist ($y$) & Dist & Thr=40 & Thr=80 & Thr=120 & Accuracy & F-score & AUC\\ \hline \hline
    ISA~\cite{bib:inferring_shared_attention} (Ex.1) & 108.5 & 85.7 & 152.7 & 8.5 & 24.9 & 48.9 & 0.41 & 0.19 & 0.41 \\ \hline
    DAVT~\cite{bib:detecting_attended} (Ex.1) & 55.6 & 26.8 & 68.2 & 58.6 & 68.5 & 79.2 & \textcolor{red}{0.52} & 0.32 & 0.58 \\ \hline
    HGTD~\cite{DBLP:conf/cvpr/TuMDGZS22} (Ex.1) & 112.5 & 65.7 & 142.7 & 20.4 & 32.9 & 46.3 & 0.18 & 0.28 & 0.50 \\ \hline
    Ours (Ex.1) & \textcolor{red}{54.3} & \textcolor{red}{26.5} & \textcolor{red}{66.5} & \textcolor{red}{59.1} & \textcolor{red}{68.7} & \textcolor{red}{79.7} & \textcolor{red}{0.52} & \textcolor{red}{0.36} & \textcolor{red}{0.64} \\ \hline \hline
    ISA~\cite{bib:inferring_shared_attention} (Ex.2) & 80.5 & 61.7 & 107.1 & 5.6 & 36.7 & 71.3 & \textcolor{red}{0.62} & 0.36 & 0.64 \\ \hline
    DAVT~\cite{bib:detecting_attended} (Ex.2) & 35.7 & 21.1 & 46.6 & 72.9 & 80.7 & 89.2 & 0.61 & 0.30 & 0.57 \\ \hline
    HGTD~\cite{DBLP:conf/cvpr/TuMDGZS22} (Ex.2) & 112.5 & 65.7 & 142.7 & 20.4 & 32.9 & 46.3 & 0.18 & 0.28 & 0.50 \\ \hline
    Ours (Ex.2) & \textcolor{red}{34.4} & \textcolor{red}{21.0} & \textcolor{red}{45.0} & \textcolor{red}{74.3} & \textcolor{red}{82.5} & \textcolor{red}{89.6} & 0.57 & \textcolor{red}{0.37} & \textcolor{red}{0.65} \\ \hline
    \end{tabular}
    \label{table:comparison_videocoatt}
\end{table*}

\begin{figure}[t]
  \begin{center}
  \includegraphics[width=\columnwidth]{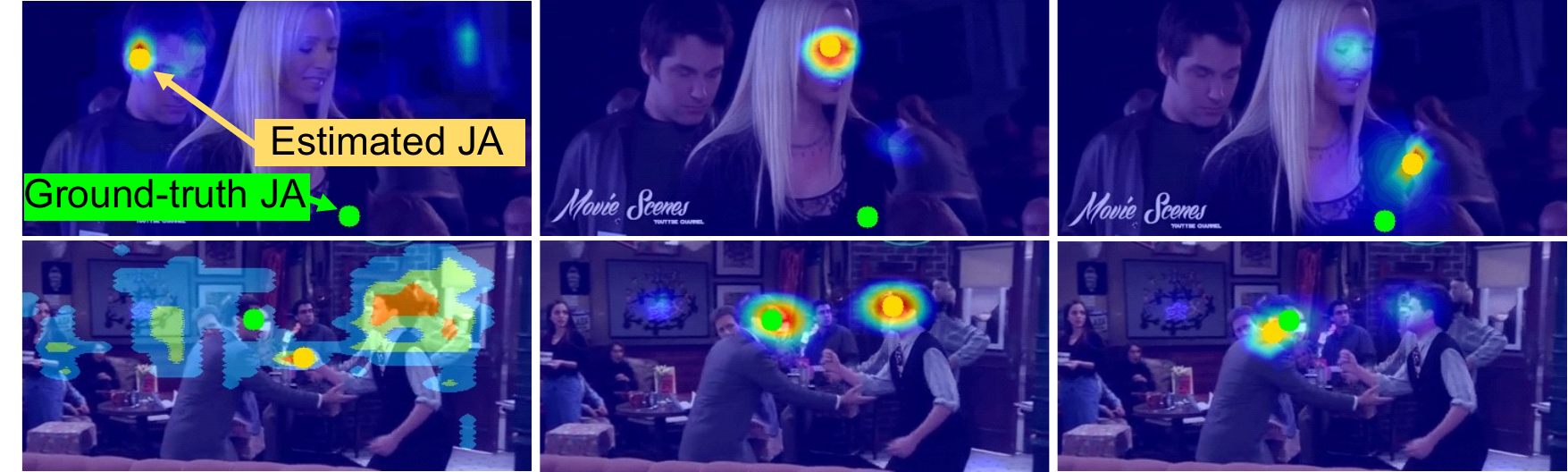}
  ISA~\cite{bib:inferring_shared_attention}\hspace{15mm}
  DAVT~\cite{bib:detecting_attended}\hspace{15mm}
  Ours~\hspace*{1mm}
  \end{center}
  \caption{Visual comparison on the VideoCoAtt dataset. See the caption of Fig.~\ref{fig:comparison_volley_previous} for details.}
  \label{fig:comparison_videocoatt_previous}
\end{figure}

\subsection{Datasets and Evaluation Metrics}
\label{subsection:exp_datasets}

The proposed method is evaluated with the following two datasets.
The Volleyball dataset, which includes many interactions among people, is mainly used to validate our contributions.
Furthermore, the VideoCoAtt dataset is used to validate the generality of our method.
%
While VideoCoAtt is used only in Sec.~\ref{subsubsection:exp_comparison_videocoatt} with Table~\ref{table:comparison_videocoatt} and Fig.~\ref{fig:comparison_videocoatt_previous}, detailed results are available in the supplementary material.

\noindent
{\bf Volleyball dataset.}
%
This dataset~\cite{bib:volleyball_dataset} has 4,830 sequences.
While each sequence has 41 frames, its center frame is annotated with the full-body bounding boxes of all players and their action classes,
each of which is either of Waiting, Setting, Digging, Falling, Spiking, Jumping, Moving, Blocking, and Standing classes.
In addition to these annotations, we newly provided the bounding box annotations of a ball whose center is regarded as the ground-truth of a joint AP.
Since the ball is not observed in Left-winpoint and Right-winpoint sequences (662 sequences in total), these sequences are omitted.
Consequently, the center frames in 4,168 sequences, consisting of 3,020 training and 1,148 test sequences~\cite{bib:volleyball_dataset}, are used in our experiments.

\noindent
{\bf VideoCoAtt dataset.}
%
This dataset~\cite{bib:inferring_shared_attention} with 380 TV-show videos is for evaluating joint APs in more general scenes.
Each frame is annotated with the bounding boxes of the ground-truth joint APs and people's heads.
The number of APs differ between frames (i.e, 0, 1, and more APs), while it is fixed to be one in all frames in the Volleyball dataset.
Since no action annotation is given to this dataset, only $\bm{l}^{i}$ and $\bm{g}^{i}$ compose $H^{i}$.
DAVT as $f^{A}$ was trained on GazeFollow~\cite{bib:gaze_follow} and VideoAttentionTarget~\cite{bib:detecting_attended} datasets.

Our experimental conditions on the aforementioned two datasets are shown in Table~\ref{table:experiments}.
In ``Att,'' people attributes used in each dataset are shown.
In each dataset, two types of experiments were conducted, namely those with the people attributes of prediction (Ex.1) and ground-truth (Ex.2).
%
%
%
%
%
In the Volleyball dataset, an annotated full-body bounding box (``Body'') is used only for head detection.
Head bounding boxes are detected in a whole image and in ground-truth full-body bounding boxes in Ex.1 and Ex.2, respectively.
In the VideoCoAtt dataset, annotated ground-truth head bounding boxes are used in Ex.2, while head bounding boxes are also detected in a whole image in Ex.1.

\noindent
{\bf Evaluation metrics.}
%
The pixel with the max value in a heatmap image is regarded as a joint AP. All results are evaluated with the Euclidean distance between the pixels of the estimated joint AP and its ground-truth. The distances along $x$ and $y$ axes are also evaluated separately for detailed analysis because AP locations are biased about $y$ axis, as mentioned in~\cite{bib:inferring_shared_attention}. In addition, the detection rate is evaluated. Each detection is considered to be successful if the distance between the estimated and ground-truth joint APs is less than each threshold. In accordance with the diameter of joint attention in images (i.e., around 30 and 40 pixels in the Volleyball and VideoCoAtt datasets, respectively), ``30, 60, and 90 pixels'' and ``40, 80, and 120 pixels'' are selected as thresholds for Volleyball and VideoCoAtt, respectively.

For the VideoCoAtt, detection accuracy is also evaluated in accordance with~\cite{bib:inferring_shared_attention,bib:detecting_attended,DBLP:conf/cvpr/TuMDGZS22}.
However, more than using accuracy is needed because there is no joint AP in over $71\%$ of test images.
To relieve the class imbalance problem, F-score and Area Under the ROC Curve (AUC) are also used. If the max value in a heatmap is greater than a certain threshold, it is regarded that a joint AP is detected. The threshold which leads to the max F-score in validation data is used.

\subsection{Training Details}
For the Volleyball dataset, the overall network consisting of ($\alpha$), ($\beta$), and ($\gamma$) in Fig.~\ref{fig:proposed_method} is trained in an end-to-end manner after pretraining ($\alpha$) and ($\beta$). For the VideoCoAtt dataset, only ($\gamma$) is trained after pretraining ($\alpha$) and ($\beta$). The learning rates for the Volleyball and VideoCoAtt datasets were 0.001 and 0.00001, respectively.

\begin{table*}[t]
    \centering
    \caption{Ablation studies in Ex.1 on the Volleyball dataset. 
    Ablated components about the people attributes
    and the network architectures 
    are separated by double lines.
    Each metric is evaluated with two results, namely $H_{JA}$ in branch ($\alpha$) and $H_{F}$ in fusion module ($\gamma$).}
    \begin{tabular}{l|r|r|r|r|r|r|r|r} \hline
    Method & Dist ($\alpha$) & Dist ($\gamma$) & Thr=30 ($\alpha$)& Thr=60 ($\alpha$) & Thr=90 ($\alpha$) & Thr=30 ($\gamma$) & Thr=60 ($\gamma$) & Thr=90 ($\gamma$)\\ \hline \hline
    Ours w/o $\bm{l}$ & 112.2 & 60.3 & 10.5 & 31.6 & 51.8 & 62.0 & 73.1 & 79.3 \\ \hline
    Ours w/o $\bm{g}$ & 138.9 & 70.8 & 22.8 & 40.7 & 53.2 & 60.5 & 72.0 & 78.3 \\ \hline
    Ours w/o $\bm{a}$ & \textcolor{red}{82.1} & 60.2 & 28.7 & 55.7 & \textcolor{red}{74.0} & \textcolor{red}{64.6} & \textcolor{red}{77.0} & \textcolor{red}{83.0} \\ \hline \hline
    Ours w/o ($\alpha$) & - & 72.0 & - & - & - & 62.0 & 72.8 & 78.6 \\ \hline
    Ours w/o ($\beta$) & 87.2 & 87.2 & 25.8 & 52.9 & 70.3 & 25.8 & 52.9 & 70.3 \\ \hline \hline
    Ours & 84.8 & \textcolor{red}{56.0} & \textcolor{red}{30.3} & \textcolor{red}{55.9} & 71.3 & 64.5 & 76.8 & \textcolor{red}{83.0} \\ \hline
    \end{tabular} 
    \label{table:ablation_volleyball_pred_atts}
    \vspace*{-2mm}
\end{table*}

\begin{table}[t]
    \centering
    \caption{Ablation studies in Ex.2 on the Volleyball dataset.}
    \begin{tabular}{l|r|r|r} \hline
    Method & Dist & Thr=30 & Thr=60 \\ \hline \hline
    Ours w/o $\bm{a}$ & 39.2 & 75.5 & 86.2 \\ \hline \hline
    Ours w/o ($\alpha$) & 77.4 & 59.7 & 69.7 \\ \hline
    Ours w/o ($\beta$) & 14.3 & 95.1 & 98.6 \\ \hline \hline
    Ours & \textcolor{red}{11.4} & \textcolor{red}{96.3} & \textcolor{red}{98.9} \\ \hline
    \end{tabular} 
    \label{table:ablation_volleyball_gt_atts}
\end{table}

\subsection{Comparative Experiments}

\subsubsection{Volleyball Dataset}
\label{subsubsection:exp_comparison_volleyball}

Our method is compared with SOTA methods~\cite{bib:inferring_shared_attention,bib:detecting_attended} on the Volleyball dataset. 
DAVT~\cite{bib:detecting_attended} is proposed as a single attention estimation, so joint attention is detected from the
mean of the independently-estimated APs of all people. 
As the visual cue of a whole image, ISA~\cite{bib:inferring_shared_attention} and DAVT~\cite{bib:detecting_attended} require a saliency map obtained by CenterNet~\cite{bib:centernet} and a raw image, respectively. 
A head location is used for cropping a head image as an input feature in DAVT~\cite{bib:detecting_attended}, while it is used only for computing the gaze direction in ISA~\cite{bib:inferring_shared_attention}.
In addition,
our method is also compared with ball detection~\cite{bib:centernet} because the ball is a strong cue for joint attention estimation in a ball game. As mentioned in Sec.~\ref{subsection:exp_datasets}, we evaluated these methods on two experimental conditions (i.e., using (Ex.1) predicted or (Ex.2) ground-truth people attributes). 

Experimental results
are shown in Table~\ref{table:comparison_volleyball}. 
Compared with all other methods, our method is better in all metrics.
Visual results are shown in Fig.~\ref{fig:comparison_volley_previous}.
While the joint attention is always on the ball in the Volleyball dataset, ball detection~\cite{bib:centernet} often fails when a ball is visually unclear (i.e., blurred).
Heatmaps estimated by SOTA methods~\cite{bib:inferring_shared_attention,bib:detecting_attended} tend to be erroneously blurred.
On the other hand, our method can successfully estimate joint attention on the ball.

\subsubsection{VideoCoAtt Dataset}
\label{subsubsection:exp_comparison_videocoatt}

%
For the evaluation with the VideoCoAtt dataset, HGTD~\cite{DBLP:conf/cvpr/TuMDGZS22} is also included as a comparative method.
HGTD is trained with GazeFollow and VideoAttentionTarget as in the original paper since it requires a ground-truth head bounding box for its training.
While HGTD is single attention estimation, joint attention is detected in a similar way as for DAVT.

Experimental results are shown in Table~\ref{table:comparison_videocoatt}. Our method is better in all metrics except the accuracy of DAVT and ISA in Ex.1 and Ex.2, respectively. It is not surprising because accuracy is optimized for F-score by validation data, as described in Sec.~\ref{subsection:exp_datasets}.



Visual results are shown in Fig.~\ref{fig:comparison_videocoatt_previous}.
In~\cite{bib:inferring_shared_attention}, the blur is worse in the bottom example.
In~\cite{bib:detecting_attended}, the estimated AP is in the right person's face, while the ground-truth AP is in the left person's face in the bottom example.
In contrast, our method can estimate the AP points more closely to their ground-truths in both examples.

\begin{figure}[t]
  \begin{center}
  \includegraphics[width=\columnwidth]{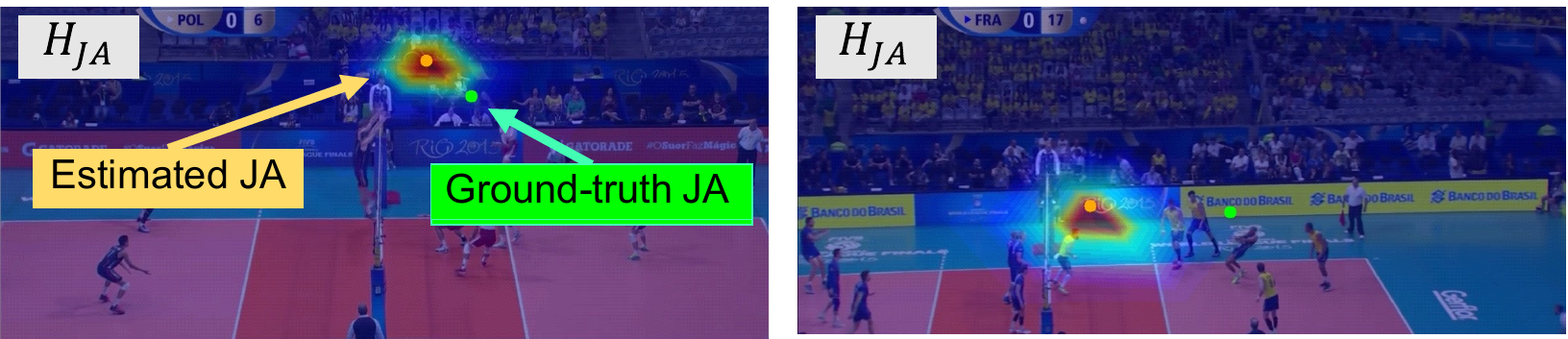}
  \includegraphics[width=\columnwidth]{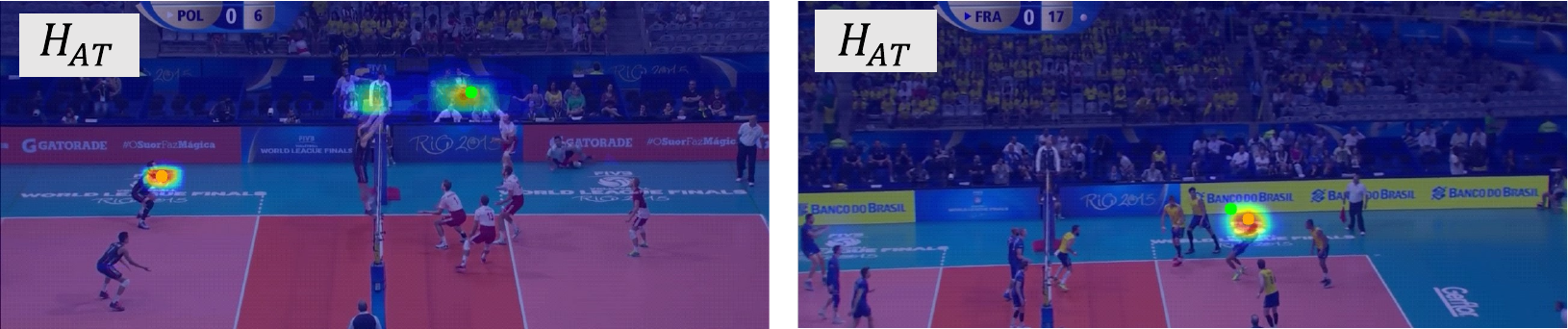}
  \includegraphics[width=\columnwidth]{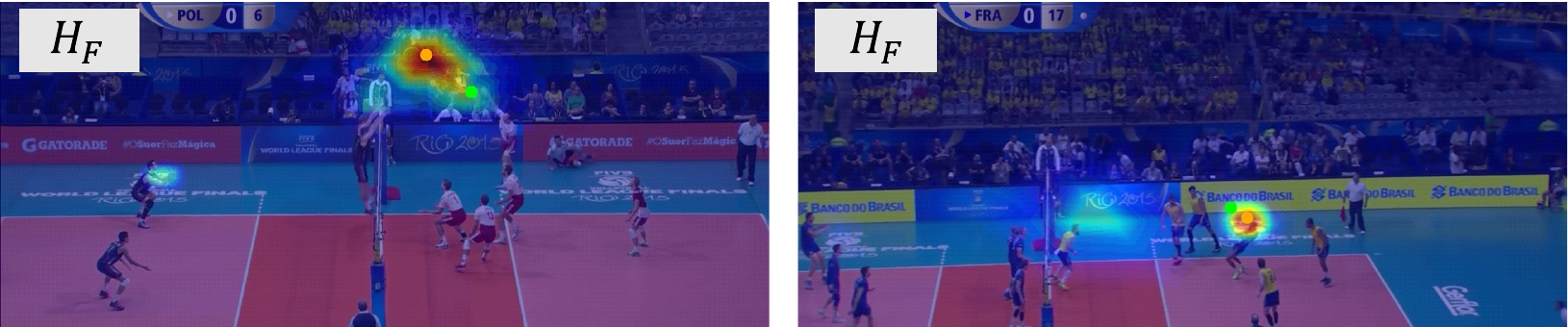}
  \end{center}
  \caption{Visualization of $H_{JA}$, $H_{AT}$, and $H_{F}$ obtained by modules ($\alpha$), ($\beta$), and ($\gamma$) in Fig.~\ref{fig:proposed_method}. $H_{F}$ is better than $H_{JA}$ and $H_{AT}$.}
  \label{fig:comparison_volley_branch}
\end{figure}

\subsection{Ablation Studies}
\label{subsection:exp_ablation}

The effect of each important component in our method is verified with the ablation studies shown in Tables~\ref{table:ablation_volleyball_pred_atts} and~\ref{table:ablation_volleyball_gt_atts} in which the results on the Volleyball dataset are shown; see the supplementary material for the VideoCoAtt dataset. We ablate either of $\bm{l}$, $\bm{g}$, and $\bm{a}$ (i.e., people attributes) by filling zero into ablated nodes in the first layer of the feature extractor network (Fig.~\ref{fig:proposed_method}). We also ablate either of network branches ($\alpha$) and ($\beta$) shown in Fig.~\ref{fig:proposed_method}. For the experiments without branch ($\alpha$) or ($\beta$), the output of each branch is regarded as the final joint attention estimation. 

In Table~\ref{table:ablation_volleyball_pred_atts}, the best results for all metrics are obtained by ``Ours'' and ``Ours w/o $\bm{a}$.''
The low performance of $\bm{a}$ can be attributed to the large recognition error of $\bm{a}$, where the accuracy of the action recognition is 53.3\%. In fact, the use of the ground-truth of $\bm{a}$ in Ex.2 significantly improves performance, as shown in Table~\ref{table:ablation_volleyball_gt_atts}.
Regarding $\bm{l}$ and $\bm{g}$, the results are improved in all metrics, as shown in Table~\ref{table:ablation_volleyball_pred_atts}. This is natural as (i) $\bm{g}$ is an essential gaze-related cue and (ii) head detection for estimating $\bm{l}$ is more reliable than the action recognition accuracy mentioned above (i.e., 53.3\%).

While the contribution of branch ($\beta$) is larger in Ex.1, that of branch ($\alpha$) is larger in Ex.2, as shown in Tables~\ref{table:ablation_volleyball_pred_atts} and \ref{table:ablation_volleyball_gt_atts}, respectively. The difference is also caused by prediction errors of $\bm{l}$, $\bm{g}$, and $\bm{a}$. The negative effect of such prediction errors is discussed in Sec.~\ref{subsection:exp_detailed}. In addition to the contribution of each branch, the combination of branches ($\alpha$) and ($\beta$) is better than the results obtained by branch ($\alpha$) or ($\beta$) independently. These results prove that each branch complementarily helps their estimation, as shown in Fig.~\ref{fig:comparison_volley_branch}.

\begin{table}[t]
    \centering
    \caption{Analysis of the negative impact caused by erroneous individual attributes, $\bm{l}$, $\bm{g}$, and $\bm{a}$, on the Volleyball dataset. GT and Pr denote the ground-truth and the prediction, respectively.}
    \begin{tabular}{l|r|r|r} \hline
    Inputs & Dist & Thr=30 & Thr=60 \\ \hline \hline
    ($\bm{l}$=GT,$\bm{g}$=GT,$\bm{a}$=GT) & 11.4 & 96.3 & 98.9 \\ \hline
    ($\bm{l}$=GT,$\bm{g}$=Pr,$\bm{a}$=GT) & 113.1 & 17.5 & 37.8 \\ \hline
    ($\bm{l}$=GT,$\bm{g}$=GT,$\bm{a}$=Pr) & 19.4 & 88.9 & 94.8 \\ \hline
    ($\bm{l}$=GT,$\bm{g}$=Pr,$\bm{a}$=Pr) & 116.2 & 16.7 & 37.5 \\ \hline
    ($\bm{l}$=Pr,$\bm{g}$=Pr,$\bm{a}$=Pr) & 150.8 & 12.4 & 26.8 \\ \hline
    \end{tabular} 
    \label{table:gt_vs_pred_analysis_gt_atts}
    \vspace*{-2mm}
\end{table}

\begin{figure}[t]
\begin{center}
\includegraphics[width=\columnwidth]{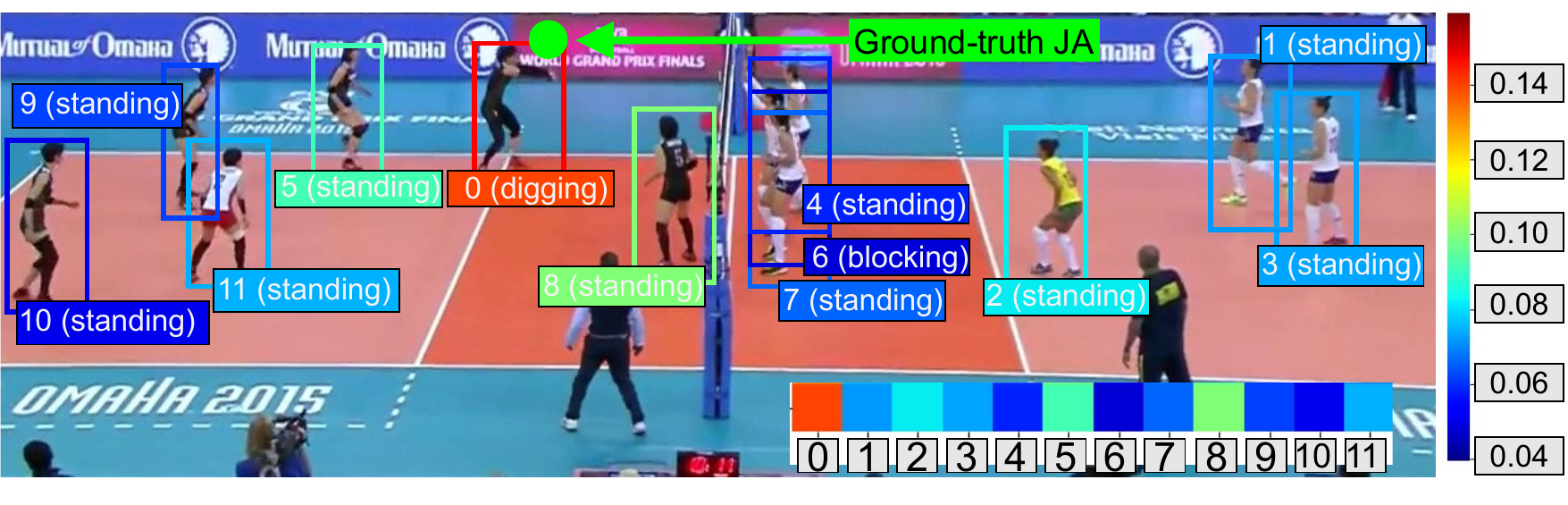}\\
\end{center}
\vspace*{-2mm}
\caption{Visualized attention values, which come from the attention map shown in Fig.~\ref{fig:self_attention}, learned by PJAT.
The values for 12 people, which are in $s_{N_{p}+1,1}, \cdots, s_{N_{p}+1,N_{p}}$ where $s_{i,j}$ denotes the $(i,j)$-th entity of the attention map, are colored and shown on the bottom right side.
The person ID ($\in 0, \cdots, 11$) is appended to this color map and each person's bounding box.}
\label{fig:self_attention_weight}
\end{figure}

\begin{figure}[t]
\begin{center}
\includegraphics[width=\columnwidth]{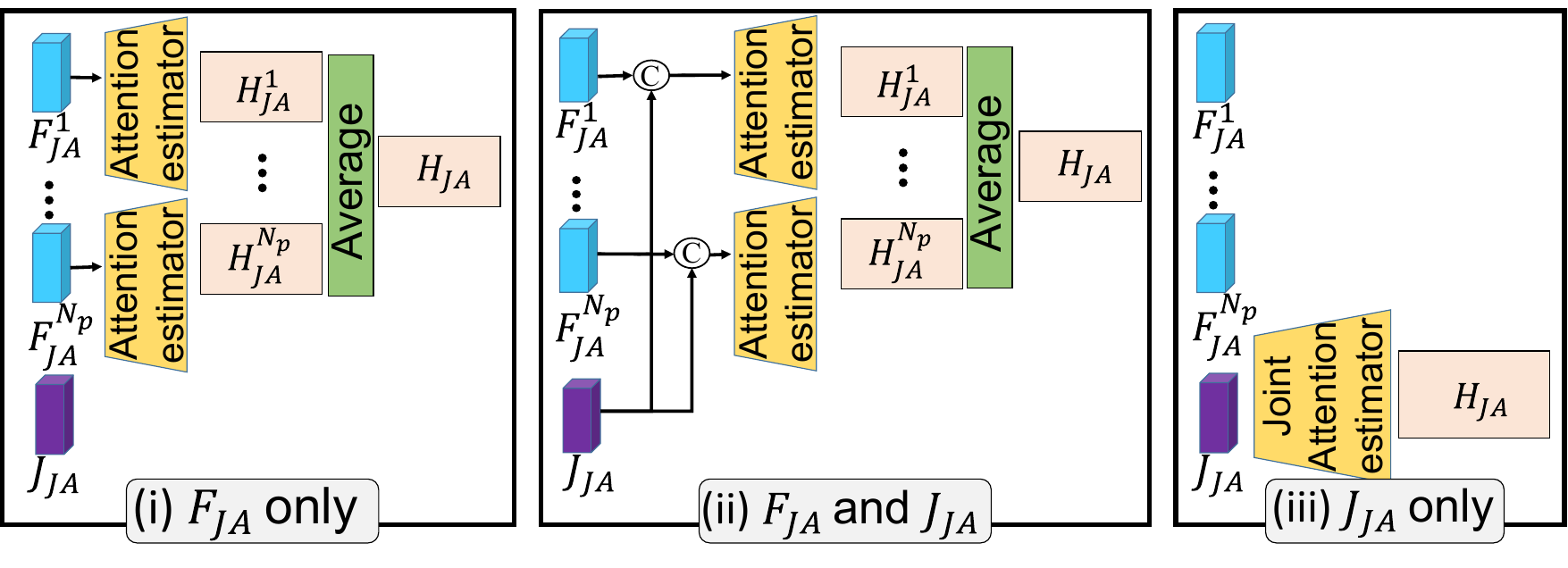}\\
\end{center}
\vspace*{-3mm}
\caption{Three types of PJAT architectures. PJAT with ``(iii) $J_{JA}$ only'' (the rightmost) is used as branch ($\alpha$) in our method.}
\label{fig:pjat_architecture}
\end{figure}

\subsection{Detailed Analysis}
\label{subsection:exp_detailed}

\noindent{\bf Negative impact of erroneous individual attributes.}
The error in the person attributes
($\bm{l}$, $\bm{g}$, and $\bm{a}$)
degrades joint attention estimation, as seen in
Tables~\ref{table:ablation_volleyball_pred_atts} and \ref{table:ablation_volleyball_gt_atts}. 
To verify the negative impact of error in each individual attribute, we use the model trained in Ex.2, but use the possible combination of ground-truths and predictions of $\bm{l}$, $\bm{g}$, and $\bm{a}$ in inference
(Table~\ref{table:gt_vs_pred_analysis_gt_atts}). Note that the three combinations with $\bm{l}$=Pr, i.e., ($\bm{l}$=Pr, $\bm{g}$=GT, $\bm{a}$=GT), ($\bm{l}$=Pr, $\bm{g}$=GT, $\bm{a}$=Pr), and ($\bm{l}$=Pr, $\bm{g}$=Pr, $\bm{a}$=GT), cannot be evaluated because the ground-truths of $\bm{g}$ and $\bm{a}$ can be used only when the location $\bm{l}$ is correct.

In comparison between ($\bm{l}$=GT, $\bm{g}$=Pr, $\bm{a}$=GT) and ($\bm{l}$=GT, $\bm{g}$=GT, $\bm{a}$=Pr), the error in $\bm{g}$ gives a much negative impact. This result is natural because the gaze direction $\bm{g}$ might be most important, while the positive effects of $\bm{l}$ and $\bm{a}$ are also validated in Tables~\ref{table:ablation_volleyball_pred_atts} and \ref{table:ablation_volleyball_gt_atts}.
The significant performance gap between ($\bm{l}$=Pr, $\bm{g}$=Pr, $\bm{a}$=Pr) and ($\bm{l}$=GT, $\bm{g}$=GT, $\bm{a}$=GT) reveals that using different GT/prediction combinations in training and test phases degrades the performance. In fact, the result of Ex.1 shown in Table~\ref{table:comparison_volleyball} is better than ($\bm{l}$=Pr, $\bm{g}$=Pr, $\bm{a}$=Pr) because the predicted attributes are used for both the training and test phases in Ex.1.


\noindent{\bf Attention values in self-attention.}
PJAT can weight the contributions of people by the self-attention mechanism,
as shown in Fig.~\ref{fig:self_attention_weight}.
It shows that the ``digging'' person (i.e., $0$-th person) closest to the AP and people nearby this person (i.e., $5$-th and $8$-th people) are highly weighted.
It can also be seen that the weights given to people doing the same action (e.g., ``standing'' of all people except $0$-th and $6$-th people) are different from each other.
This is evidence that PJAT can learn complex people interactions so that the weight of each action is not fixed but changes depending on other attributes such as the location and gaze direction.

\noindent{\bf PJAT architecture comparison.}
Table~\ref{table:ablation_volleyball_pjat} shows the comparison of different PJAT architectures in Fig.~\ref{fig:pjat_architecture}. To estimate $H_{JA}$, ``$F_{JA}$ only'' takes the average of $H_{JA}^{i}$, the AP heatmaps estimated from $\bm{F}_{JA}^{i}$. 
In ``$F_{JA}$ and $J_{JA}$'', $\bm{J}_{JA}$ is also used to estimate each $H_{JA}^{i}$. 
While the two methods aggregate estimated maps $H_{JA}^{i}$ to compute $H_{JA}$, ``$J_{JA}$ only'' estimates $H_{JA}$ directly from $\bm{J}_{JA}$. 
``$J_{JA}$ only'' shows the best performance as it can directly utilize person-level contribution weights. However, the gap from ``$F_{JA}$ only'' is small, which may reflect the characteristics of the Volleyball dataset, i.e., the large number of people in a scene yields robust estimation with simple averaging.

\noindent{\bf Pixelwise vs. imagewise.}
Pixelwise estimation with PJAT is compared with general imagewise heatmapping. As shown in Table~\ref{table:ablation_volleyball_pjat}, ``Ours'' outperforms ``$J_{JA}$ only for imagewise'' because pixelwise estimation with positional information avoids an ill-posed problem mentioned in Sec.~\ref{section:introduction}.

\noindent{\bf Fusion module comparison.}
Three fusion modules are compared in Ex.1. ``CNN'' fuses $H_{JA}$ and $H_{AT}$ by convolutional layers, where the details of the architecture are shown in the supplementary material. ``Average'' takes the average of $H_{JA}$ and $H_{AT}$ to compute $H_{F}$. In ``Weighted,'' the weight coefficients for $H_{JA}$ and $H_{AT}$ (i.e., $W_{JA}$ and $W_{AT}$ in Sec.~\ref{subsection:fusion}) are optimized.
As shown in Table~\ref{table:ablation_volleyball_fusion_module}, 
``CNN'' is worse than the others due to inefficient convolution for the sparse heatmaps, $H_{JA}$ and $H_{AT}$. While the gap from ``Average'' is small, ``Weighted'' requires only a few parameters, leading to stable weight estimation and better results.

\begin{table}[t]
    \centering
    \caption{Comparison of different heatmap generators in branch ($\alpha$) in Ex.1 on the Volleyball dataset. Ours uses ``(iii) $J_{JA}$ only'' for pixelwise estimation. See the supplementary material for Ex.2 and the results on VideoCoAtt.}
    \begin{tabular}{l|r|r|r} \hline
    Method & Dist ($\alpha$) & Thr=30 & Thr=60 \\ \hline \hline
    (i) $F_{JA}$ only & 87.8 & 25.9 & 51.6 \\ \hline
    (ii) $F_{JA}$ and $J_{JA}$ & 93.7 & 25.2 & 50.3 \\ \hline \hline
    $J_{JA}$ only for imagewise & 126.7 & 10.7 & 28.4 \\ \hline \hline
    (iii) $J_{JA}$ only (Ours) & \textcolor{red}{87.5} & \textcolor{red}{27.1} & \textcolor{red}{53.4} \\ \hline  
    \end{tabular}
    \label{table:ablation_volleyball_pjat}
\end{table}

\begin{table}[t]
    \centering
    \caption{Comparison of different fusion modules in Ex.1 on the Volleyball dataset. See the supplementary material for Ex.2 and the results on VideoCoAtt.}
    \begin{tabular}{l|r|r|r} \hline
    Fusion & Dist & Thr=30 & Thr=60 \\ \hline \hline
    CNN & 94.6 & 44.1 & 68.8 \\ \hline
    Average & 58.1 & 61.8 & 76.2 \\ \hline \hline
    Weighted (Ours) & \textcolor{red}{56.0} & \textcolor{red}{64.5} & \textcolor{red}{76.8} \\ \hline 
    \end{tabular}
    \label{table:ablation_volleyball_fusion_module}
\end{table}


\section{Concluding Remarks}

We addressed joint attention estimation by modeling the interaction of people attributes
as rich contextual cues.
To relieve the difficulty in estimating a high-dimensional heatmap from a low-dimensional latent vector, we proposed the Position-embedded Joint Attention Transformer (PJAT).
Our method achieves state-of-the-art results on two significantly-different datasets, which prove the wide applicability of our method.
This paper focused on the image domain, as with~\cite{DBLP:conf/wacv/SumerGTK20,DBLP:conf/cvpr/TuMDGZS22}, to validate our key idea (i.e., activity- and interaction-aware joint attention heatmapping) more clearly, which is directly applicable also to videos, as with~\cite{bib:detecting_attended,bib:inferring_shared_attention}.
On the other hand, the use of video-specific features is important future work.

{\small
\bibliographystyle{ieee_fullname}
\bibliography{egbib}
}

\end{document}